\documentclass[runningheads]{llncs}
\usepackage{graphicx}
\usepackage{caption}
\usepackage{subcaption}
\usepackage{cite}
\begin{document}
\title{Artificial Open World for Evaluating AGI: a Conceptual Design}
%
%
\author{Bowen Xu\inst{1, 2}\orcidID{0000-0002-9475-9434} \and
Quansheng Ren\inst{1,2}\orcidID{0000-0001-8698-9603}
}
\authorrunning{B. Xu, and Q. Ren}
%
\institute{School of Electronics, Peking University, Beijing, 100871, China \and
\email{\{xubowen, qsren\}@pku.edu.cn}
}
\maketitle              
\begin{abstract}
    How to evaluate Artificial General Intelligence (AGI) is a critical problem that is discussed and unsolved for a long period. In the research of narrow AI, this seems not a severe problem, since researchers in that field focus on some specific problems as well as one or some aspects of cognition, and the criteria for evaluation are explicitly defined. 
    By contrast, an AGI agent should solve problems that are never-encountered by both agents and developers. However, once a developer tests and debugs the agent with a problem, the never-encountered problem becomes the encountered problem, as a result, the problem is solved by the developers to some extent, exploiting their experience, rather than the agents. This conflict, as we call \textit{the trap of developers' experience}, leads to that this kind of problems is probably hard to become an acknowledged criterion. 
    In this paper, we propose an evaluation method named Artificial Open World, aiming to jump out of the trap. The intuition is that most of the experience in the actual world should not be necessary to be applied to the artificial world, and the world should be open in some sense, such that developers are unable to perceive the world and solve problems by themselves before testing, though after that they are allowed to check all the data. The world is generated in a similar way as the actual world, and a general form of problems is proposed. A metric is proposed aiming to quantify the progress of research. 
    This paper describes the conceptual design of the Artificial Open World, though the formalization and the implementation are left to the future.

\keywords{Evaluation \and Artificial Open World \and Artificial General Intelligence}
\end{abstract}
\section{Introduction}
    
In AGI research, the problem, of how to evaluate AGI, \textit{itself} is a problem. In ``narrow AI\cite{goertzel2007artificial, wang2012theoretical}'', this seems not a severe problem, since in that field the criteria are explicit, for example, in the field of Image Recognition, researchers aim to rise up the accuracy of classification and use any tricks to solve that problem. Few may deny that datasets, as problems for evaluation, play an important role in the rapid progress of narrow AI. 
However, in AGI research, it is quite a different story on evaluation. Despite different definitions, goals, and pathways of AGI\cite{wang2012theoretical}, under the perspectives of intelligence in Sec. \ref{sec:def}, we hold that an AGI agent should solve problems that are unknown to both agents and developers. However, once a developer tests and debugs the agent with a problem, the unknown problem becomes a known problem, as a result, that problem is no longer suitable for evaluating AGI agents -- the developers are able to construct a problem-specific system that could not be applied to other situations, and the performance of a system in this problem does not reflect the progress on AGI. We call this trouble \textit{the trap of developers' experience}. To deal with this trouble, an alternative is to design new problems constantly\cite{genesereth2013international}, though we adopt a different path to jump out of the trap in this paper, \textit{i.e.}, designing an artificial world.
The Artificial Open World is generated in a similar way as the actual world, currently based on a classical world-outlook. The world should be open, in the sense that the causations in the world are time-varying on some abstract level, and problems to be solved are continuously changing. Implicitly infinite instances of the world can be generated so that for any of the instances, developers are possibly unable to perceive the world and solve problems by themselves based on their experience of the actual world. Nevertheless, after testing, developers are allowed to check all the data and analyze the activities of agents, and then perceive the instance of the world. The developers' knowledge of one instance of the world is not necessary to be applied to another instance, such that facing a new instance, an agent has to solve problems by exploiting its own intelligence. The world should be generated in a similar way as the actual world, so that the knowledge of the generation is allowed to be known by agents in advance, because the agent with the knowledge would be still able to adapt to the actual world, without being disturbed by problem-specific knowledge from developers.
To quantify the progress of AGI research, a metric is also proposed. We consider three aspects of performance, \textit{i.e.}, the speediness of adaptation, the goodness of adaptation, and the goodness of generalization (see Sec. \ref{sec:metric}), and they should be merged together into one value, as the measure of intelligence. It should be noted that the value is a lower-bound of intelligence, and complicated situations partially stem from the competition between different agents in the world.

\section{What Intelligence is}
\label{sec:def}

Before proposing the evaluation method, in this section, we should first figure out what \textit{that thing} which is called \textit{intelligence} is. We are not trying to propose a definition of intelligence within a brief sentence, but we are trying to describe our perspectives on that thing which is called intelligence.

Different perspectives on intelligence lead to different work. If one regards intelligence as the ability to solve complex problems, he or she would specify a sufficiently complex problem to be solved by a machine \cite{campbell2002deep, silver2017mastering}. If one treats intelligence as a set of cognitive functions, he or she would model human cognition with a cognitive architechture\cite{hart2008opencog} or would let machines have capabilities that are presented in human beings, such as image recognition, natural language processing, \textit{etc}. However, an agent, which possesses that thing which is called intelligence, should not merely solve several specified problems, no matter how complex they are, and should not has only parts of the capabilities of human cognition. Therefore, to distinguish the goal of creating a general-purpose system and the specific methods of solving specified problems, the term \textit{AGI (Artificial General Intelligence)} is invented \cite{goertzel2007artificial}. An AGI agent should own that thing which is called intelligence. What is that thing after all?

The definitions of intelligence is discussed by a lot of predecessors(\textit{e.g.}, \cite{legg2007collection, goertzel2014artificial, wang2020defining}). Among the definitions, Pei Wang's grasped some essential aspects of intelligence. In Wang's definition, \textit{“Intelligence is the capacity of an information-processing system to adapt to its environment while operating with insufficient knowledge and resources”}\cite{wang2020defining, wang1995non}, where \textit{insufficient knowledge and resources} means being \textit{finite}, being \textit{open}, and working in \textit{real-time}. Being \textit{finite} means a system has insufficient spatial resources to store information and insufficient time to process information. As an intuition, an algorithm which searches exhaustively an answer, which is stored in an infinte memory, is not of intelligence. In this sense, insufficiency is critical. Being \textit{open}, in Wang's theory, means the content of tasks should not be specified before the system has been developed. Working in \textit{real-time} means multiple tasks may occur in the same time, and one task may interrupt another. Adaptation in the definition refers to “the mechanism for a system to summarize its past experience to predict the future situations accordingly, and to allocate its bounded resources to meet the unbounded demands”. In Pei Wang's theory \cite{wang2013non}, the constraints of insufficient knowledge and resources have been placed at the forefront, though they are obvious in human beings' and machines' lives. 

François Chollet proposed the “generalization spectrum” -- \textit{absence of generalization}, \textit{local generalization}, \textit{broad generalization}, and \textit{extreme generalization} -- and use the word intelligence to refer to the \textit{extreme generalization}\cite{chollet2019measure}. An agent, \textit{e.g}. a sorting algorithm, with absence of generalization can only handle those situations with no uncertainty. An agent, \textit{e.g.} current machine learning systems, with local generalization, should handle a single task or a few tasks, which are well scoped by developers. An agent, with broad generalization, should generalize to unknown unknowns across a broad category of related tasks, for example, an image classifer could recognize dog while it is trained with cat images. An agent, with intelligence, as Chollet considered, should generalize to unknown unknowns across an unknown range of tasks and domains. Chollet may presuppose implicitly that unknowns and knowns have similarity on some abstract level, and an agent who is able to identify that kind of similarity is of intelligence. We generally agree Chollet's view that an agent with intelligence should adapt to “unknown unknowns across an unknown range of tasks and domains”, though the meaning of “adapt” here may not be the same as that in Chollet's definition, the meaning of which we approve is closer to that in Wang's.

As our position, we hold that intelligence is a unity, which implies that it is a whole which can be described from different points of view. From one perspective, intelligence is a property with which an agent is able to deal with tasks in an open environment with limited resources. From another perspective, intelligence is an object which involves principles of representation-interaction. Informally, an environment is open, which means that causations in the environment is time-varying to some extent. 

As a further illustration, facing with the open environment, on one hand, an agent with intelligence should generalize to unknowns scenes, which means that, facing problems which are not encouterd before, the agent should take reasonable solutions based on its past experience. The agent have to use the similarity on some abstract level to deal with the unknowns. On the other hand, after encoutering a series of similar problems, which are expected to be solved by the agent, it should adapt to the problems as quickly as possible and performs as well as possible. Furthur more, as an explicit claim, the agent should be able to match a special-purpose system designed for specific tasks without losing the ability to adapt to new problems. Intelligence is the thing which facilitates an agent to meet the requirements mentioned above.

\section{Evaluation}


It is merely impossible to exhaustively review plenty of proposals and work on evaluating intelligence in this paper. Nevertheless, we briefly review some pieces of work and then propose our solution.
A typical sort of evaluation is similar to I-athlon (Olympic Decathlon of Intelligence)\cite{Adams16Iathlon}: a series of cognitive tasks are defined to test different capabilities. Broadly speaking, that evaluation method seems to assume that the more tasks an agent can fulfill and the better the agent performs in a task, the more intelligent the agent is. Some work focuses on the difficulty of problems and designs some puzzles to be solved by agents, \textit{e.g.}, the Bongard problem\cite{GEB99}. To evaluate cognitive architecture, some metrics are proposed, \textit{e.g.} \cite{wray2007metrics}, and we agree with some of them, especially the metric ``taskability'', which is the ability to adapt to new tasks. To evaluate human-level AGI, Goertzel and Bugaj proposed to build a school environment and educate agents in it, and whether an agent has some skills, \textit{e.g.}, logical-math, music, story understanding, \textit{etc.}, determines the extent of intelligence\cite{goertzel2009agi}. Regardless of the feasibility in practice, there is a more severe problem: as Wang pointed out,

\begin{quotation}
    Though such activities do stimulate interesting research, it still has the danger of leading the research to problem-specific solutions, no matter how carefully the problems are selected — after all, this was why problems like theorem proving and game playing were selected in the early days of AI, and the resulting techniques have not been generalized to other fields very well. \cite{wang2010evaluation}
\end{quotation}

\subsection{The Trap of Developer's Experience}
\label{sec:trap}

Those of AGI evaluation also encountered the same trouble as those work on evaluating narrow AI, \textit{e.g.} datasets such as ImageNet\cite{deng2009imagenet}, games such as Chess\cite{campbell2002deep} and Go\cite{silver2017mastering}, \textit{etc.}: developers may solve the problem and exploit their problem-specific knowledge, using any tricks, to program an agent. The problems for evaluation are hard to avoid this kind of cheating, such that a problem-specific method performs better than a general system, which makes the problems unsuitable to evaluate a general system. 

Even if at first a problem is not permited to be seen by developers, after testing an agent, the problem should be presented to the developers for further analysis, otherwise, this kind of problems is almost not suitable for advancing the research. As thus, the unknown problem becomes a known problem, and developers' experience on the specific problems would inevitably impact their designing the model of intelligence. 

The agent with intelligence is necessary to adapt to an open environment, as we claim in Sec. \ref{sec:def}. The environment could be complex or simple, actual or artificial, however, openness plays a critical role. The environment human faces is an actual, complex, and open one. The environment AlphaGo\cite{silver2017mastering} faces is an artificial, simple, and closed one. The environment of ImageNet\cite{deng2009imagenet} is an actual, complex, and closed one. If the environment is closed, which means that the problems can be one by one solved by human developers, it is almost inevitable for developers to introduce their problem-specific experience to the machine. Eventually, it is not a machine but a human who solves problems.

This trouble, which is the reason why the traditional problems have the danger of leading the research to problem-specific solutions, is what we call \textit{the trap of developer's experience}.

What we need is an admitted criterion, which could be used to compare different AGI agents within a relatively long period. To jump out of \textit{the trap of developers' experience}, we first consider some overall principles of designing the evaluation method and then give some more detailed description of our proposal, the \textit{Artificial Open World}.

\subsection{Overall Principles}


An AGI agent is required to be adaptive when faced with various problem in an open environment, and to find reasonable solutions without adaptation facing with new circumstances; simultaneously, for a specific problem, the agent should perform well with sufficient training, while it is still able to adapt to other problems and environments. Therefore, we should test how fast and how well an agent adapt to new environments and how well the agent generalize to new environments which are similar to the past.

Further, we suggest several criteria, for designing the Artificial Open World, that an AGI test should follow:
\begingroup
\renewcommand\labelenumi{(\theenumi)}
\begin{enumerate}
    \item \textbf{Independence.} The test should be abstract and independent of the actual world, 
    which means that developers' experience of the actual world is not necessary to be applied to the artificial world.
    When solving problems in such a world, there are no problem-specific priors of developers, because the developers and the agents live in two worlds independent of each other -- for example, knowledge of vision in the actual world does not have to be true in the artificial world.  
    \item \textbf{Similarity.} The artificial world is similar with the actual world in the process of generation, \textit{i.e.}, the actual world is similar to one instance of the artificial world.
    If an agent performs well in the artificial world, it will be adaptive not only to our humans' actual world, but also to any worlds which has common natures in some sense.
    The knowledge about the generation is permited to be priori knowledge of developers, since even though a developer convert this knowledge into a skill of an agent, the agent is still able to adapt to the actual world.
    \item \textbf{Openness.} The world should be open in the sense that causations in the world are time-variying to some extent, and \textit{new} problems can be generated continuously.
    \item \textbf{Asymmetry.} To generate the world is easy, but to conjecture directly the parameters or structures of the world inversely should be hard or even impossible, so that developers cannot use the artificial-world-specific algorithm to acquire knowledge, which is only applied to one instance of the world.
\end{enumerate}
\endgroup
\noindent

\subsection{Conceptual Design of Artificial Open World}\label{sec:AOW}

\begin{figure}[b!]
    \begin{center}
    \includegraphics[width=1.0\linewidth]{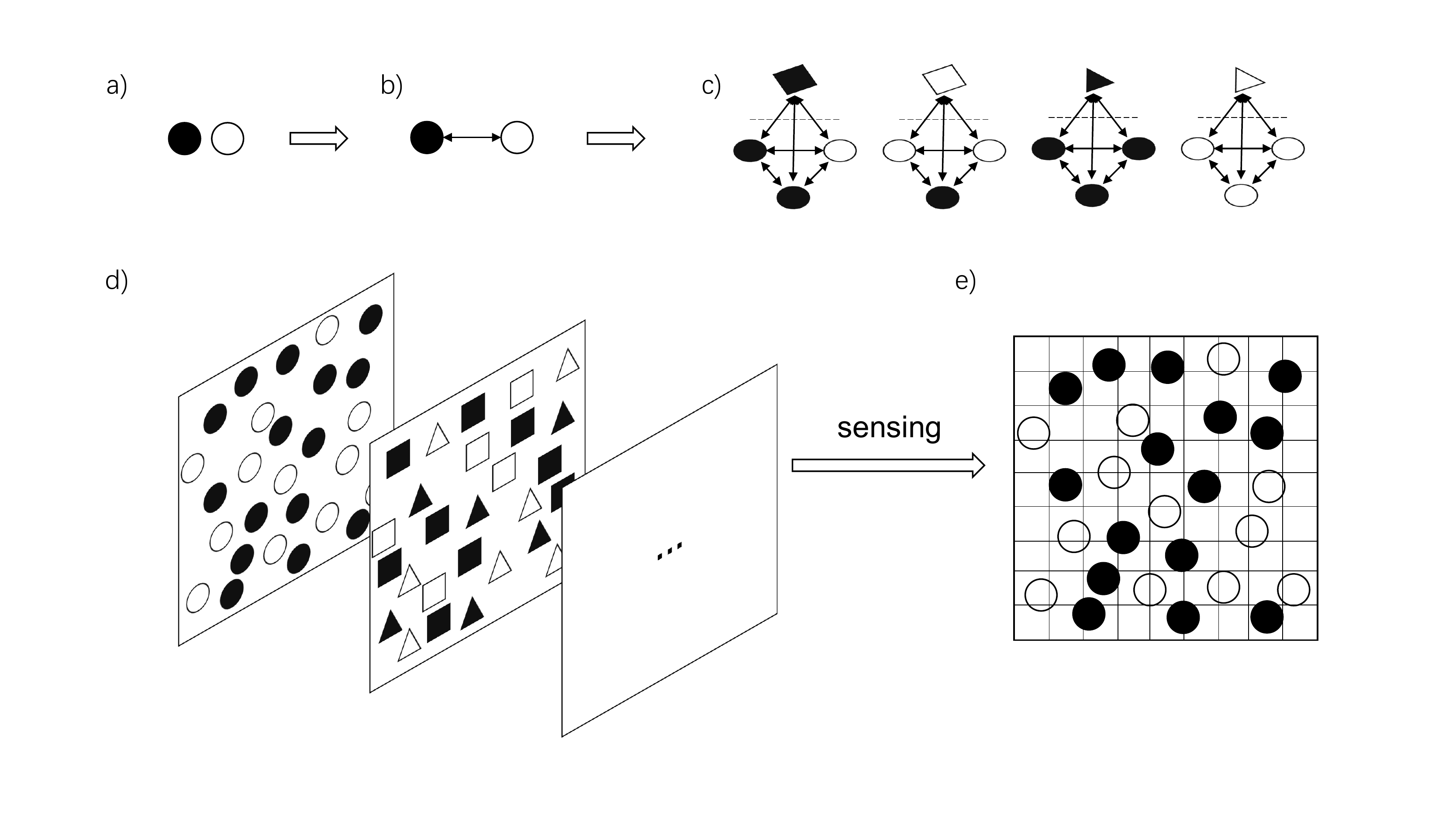}
    \end{center}
    \caption{The generation process of the Artificial Open World.} \label{fig:generation}
\end{figure}

\subsubsection{Generation.}

There are three steps to generating the world. The first step is \textit{differentiation}. As shown in Fig. \ref{fig:generation}a, two different kinds of \textit{entities} are generated: one is positive, and another is negative. A number of entities are generated in the world, and the basic property of an entity is its spatial position. The second step is generating \textit{causations}, as shown in Fig. \ref{fig:generation}b and Fig. \ref{fig:generation}c. Every two entities interact with each other, and several entities combine together as a whole, the whole as an entity interacts with others. A relation of the interaction is called a \textit{causation}. Through the combination, the world is hierarchical, as shown in Fig. \ref{fig:generation}d. The third step is to import the mind. The entities with a mind constitute an agent, and the entities themselves constitute the body of the mind. There is also a set of causations, as an interface, between the mind and the body. The world without the mind is mechanical, rigid, and inanimate, however, the mind makes the world complex and vibrant -- just as in a board of Go, players' mind leads to various complex situations. 

The causations should be generated in some way. For example, the causation between two entities were a second-order differential equation, and the coefficients in the equation were randomly generated; further, the equation were not necessary to be a second-order differential equation, and the form of the equation were randomly generated. The causations do not have to be the same as those in the actual world so that the developers' experience of the actual world is almost unsuitable to the generated world.

Some of the causations are stable, while some of the causations are time-varying. For example, in the hierarchical structures shown in Fig. \ref{fig:generation}d, the causations in lower levels are fixed, and the ones in higher levels are continuously changing. This is similar to the actual world: the dynamics of microscopic particles are stable, while the weather of an area is changing.

\subsubsection{Mind-Body Interface.}

Here, we use the term \textit{mind} to refer to \textit{intelligence}, though they are not completely equivalent in some sense\cite{xu2021gap}.
To enable an agent with the mind act and solve problems in the world, there are two kinds of mind-body interfaces, \textit{i.e.}, sensor and motor. The former is a causation where the cause is the entities outside the mind, and the latter is a causation where the cause is the mind.
Through sensor, the mind can sense the basic entities (in Fig. \ref{fig:generation}a), however, due to the limitation of resources, the data sensed is a projection of the entities, with a certain resolution, as shown in Fig. \ref{fig:generation}e. For example, the retina cannot sense every atom accurately but can sense the environment with a certain resolution. Through motor, the mind can affect spatial positions of the body, which is a set of entities, and through further interactions between entities, the mind can affect a broader range of the world.

At the current stage, there are two considerations on the body. One is that the mind has a fixed body, which would not be destroyed by the environment so that the mind can survive in the world to solve problems for further evaluation. The other is that the body is evolved and could be destroyed, and one goal of the mind is to maintain the existence of the body.

\subsubsection{Problems.}

To measure the intelligence, we should define the problems to be solved in the world. However, once a specific problem is defined, developers would solve the problem and put the skills into a machine, as a result, it is not machines but developers who solve problems. To avoid the \textit{trap of the developers' experience}, we consider a general form of the problems. 

The objective status of the entities at time $t$ is denoted as $s_t$, and the target status at time $t$ is denoted as $s_t'$. A problem is defined as the pair $(s_t, s_t')$. To solve the problem is defined as to find a series of actions, which is denoted as $a$, so that $s_t$ is evolved to $s_t'$. An agent is informed of the problem in a certain way and gave a score for solving the problem. The considerations for calculating the score are illustrated in the next sub-section \textit{Metric}.

As thus, the implicitly countless problems could be generated, even if a developer debugs the program and checks how an agent solves a problem, the future encountered problems are not solved by the developer, and the developer's experience is not necessarily suitable for those cases.

\subsubsection{Metric.}
\label{sec:metric}

To evaluate the adaptability of an agent, an intuition is that the agent should solve the problems with fewer observations and attempts, simultaneously, for a problem which are similar, on some abstract level, to those solved ones, an agent should solve the problem, to some extent, without attempts, \textit{i.e.}, the agent should generalize its experience to the new problem.

Based on these considerations, there are some indicators to be measured. We denote the number of observations for an agent to solve a problem as $O$ and the duration consumed as $D$. The indicators $O$ and $D$ are objective in the sense that they are independent of the implementation of agents. We denote the memory resources consumed for an agent to solve a problem as $M$ and the calculation resources consumed as $C$. The indicators $M$ and $C$ are subjective in the sense that they depend on the implementation, \textit{e.g.}, programming language, hardware, theoretical model, \textit{etc}. Whenever an agent solves a problem, it obtains a score, denoted as $S$. The score $S$ should be negatively related to $O$ and $D$. The score $S$ can be normalized by $M$ and $C$ so that different AGI models can be Relatively fairly compared.

\begin{figure}[htb]
    \centering
    \begin{subfigure}{0.49\textwidth}
        \centering
        \includegraphics[width=0.8\linewidth]{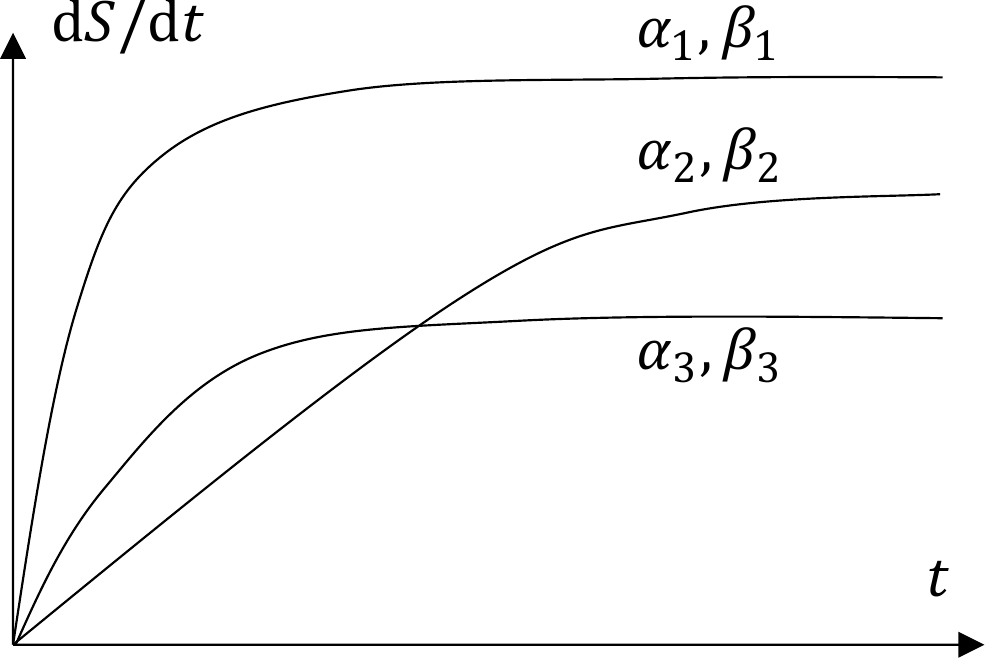}
        \caption{}
        \label{fig:adaptation-curves-sub1}
    \end{subfigure}   
    \begin{subfigure}{0.49\textwidth}
        \centering   
        \includegraphics[width=0.8\linewidth]{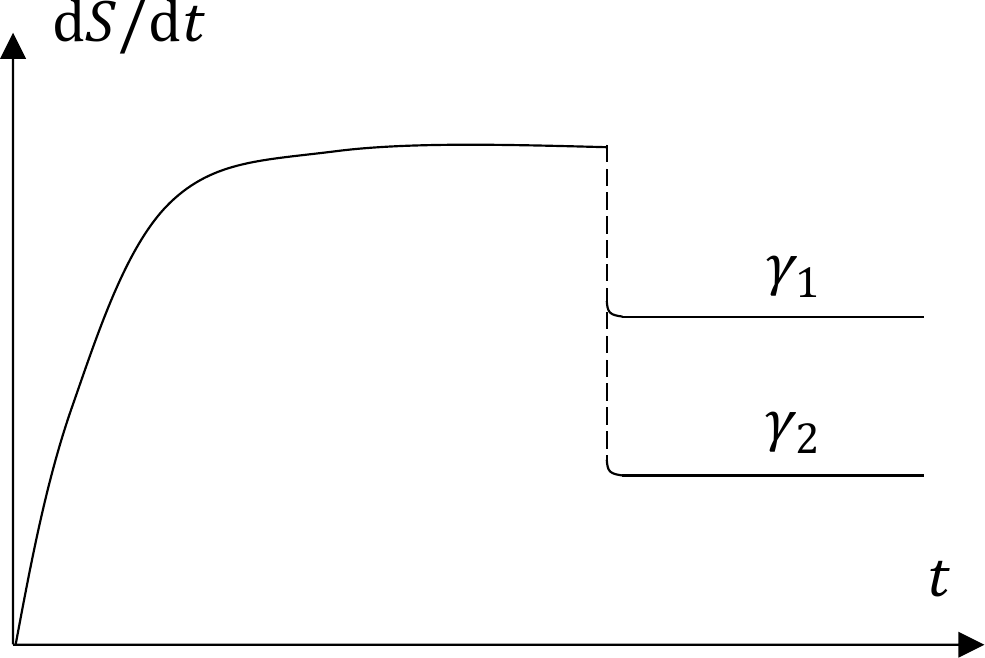}
        \caption{}
        \label{fig:adaptation-curves-sub2}
    \end{subfigure}
    \caption{Curves for evaluation. (a) Comparison on adaptation, where $\alpha$ indicates the speediness of adaptaion, and $\beta$ indicates the goodness of adaptaion. In this figure, $\alpha_1 > \alpha_3 > \alpha_2$ and $\beta_1>\beta_2>\beta_3$. (b) Comparison on generalization, where $\gamma$ indicates the goodness of generalization. In this figure, $\gamma_1 > \gamma_2$.}
    \label{fig:adaptation-curves}
\end{figure}

Given the scores which varies with time, the time derivative of $S$, $\texttt{d}S/\texttt{d}t$, is calculated, and the typical curves are drawn in Fig. \ref{fig:adaptation-curves-sub1}. The derivative $\texttt{d}S/\texttt{d}t$ reflects the performance of adaptation to some extent. The faster a curve rises up, the faster the agent adapts to a new circumstance, \textit{i.e.} it reflects the speediness of adaptation; the higher a curve reach, the better the agent adapts to a particular circumstance, \textit{i.e.} it reflects the goodness of adaptation. After the causations are changed, the performance of the agent to obtain the scores would drawdown, and the extent of it reflects the goodness of generalization. In some way, indicators $\alpha$, which denotes the speediness of adaptation, $\beta$, which denotes the goodness of adaptation, and $\gamma$, which denotes the goodness of generalization, are calculated all based on $\texttt{d}S/\texttt{d}t$.
Finally, there should be an overall metric of intelligence, $I=M(\alpha, \beta, \gamma)$, where $M$ is a function to merge the three indicators into one value $I$.

We argue that the metric $I$ is a lower-bound of the measure: an agent is voluntary in some sense, which means that it may choose to do nothing at all in the world, without presenting its wisdom. Nonetheless, in a test, to increase the lower-bound, developers are allowed to modify some parameters of their models, so that agents are proactive in solving problems. In this sense, the metric $I$ provides evidence that an agent is of intelligence.

We argue that there would be two stages of evaluating AGI. At the first stage, an agent is tested in the artificial world without other agents participating in; thus, at this stage, the metric $I$ is an absolute one, which only reflects the ability of understanding the world. At the second stage, multiple agents lives in the same world, and more complex phenomena would emerge. Agents would compete and cooperate with each other, and communicate with each other, when game behaviors and language might emerge; thus, at this stage, the metric $I$ is a relative one, and those agents who is better at game, or has the capability of language, might obtain relatively higher $I$.



\subsubsection{Future Work.} 

We will formalize the description in Sec. \ref{sec:AOW} and implement the Artificial Open World in the future so that researchers can easily install the environment, test their agents, and compare their models with others' practically.

There are still some theoretical troubles of \textit{Metric} in Sec. \ref{sec:metric}. For example, how to quantify the subjective indicators $M$ and $C$ in practice, and how to adjust the indicators $O$ and $D$ according to the difficulty of $s_t$ reaching $s_t'$ without agents' efforts.

The previous problems, including Game of Go, theorem proving, image recognition, natural language understanding, \textit{etc}, should be special cases of the problems in the artificial world, however, this deserves further justification. The issue of causation, which is an important concept in our design, is discussed for a long time in philosophy\cite{schrenk2016metaphysics}, as well as in AGI\cite{wang2015issues}, and the term causation should be further clarified.

\section{Discussion}

An interesting issue is the logic in the Artificial Open World. The logic can be adaptive, which means that the logic rules and their truth functions are acquired through interactions with the world, but are not designed and fixed. More concretely, for example, in Non-Axiomatic Logic, an acquired relation is represented as $\langle (*, T_1, T_2) \rightarrow R \rangle $, where $R$ is the relation term; a syllogistic rule can be $\langle (\wedge, \langle (*, \$M, \$P) \rightarrow inheritance \rangle, \langle (*, \$S, \$M) \rightarrow inheritance \rangle) \Rightarrow \langle (*, \$S, \$P) \rightarrow inheritance \rangle \rangle $. Suppose that the truth value of the two premises $\langle (*, \$M, \$P) \rightarrow inheritance \rangle$ and $ \langle (*, \$S, \$M) \rightarrow inheritance \rangle$ are $(f_1, c_1)$ and $(f_2, c_2)$, and that the truth value of the conclusion $\langle (*, \$S, \$P) \rightarrow inheritance \rangle$ is $(f, c)$. The truth value $(f, c)$ is determined by $(f_1, c_1)$ and $(f_2, c_2)$ through a function $F(f_1, c_1, f_2, c_2)$. The function $F(\cdot)$ is acquired via experience, rather than identified in advance. The intriguing questions occur: will the agent in the artificial open world follow the same logic which is discovered in the actual world? Will the logics, which are learned by agents in different configurations of the world, be the same to some extent? Will the logics emerged be appropriate for the agent in the actual world? If the answers are ``yes'', it will be quite strange that the logic seems a universal existence. If the answers are ``no'', then the artificial open world puts forward a higher demand for researchers to design an adaptive logic.

\section{Contributions\&Acknowledgements}

Bowen Xu proposes the main idea and writes this paper; Quansheng Ren, who reviews and modifies the paper, points out the key idea that the complexity of the world stems from agents' behaviors. We thank Pei Wang for sharing some pieces of literature on evaluating AGI. We thank those who review this paper.


%
%
%
\bibliographystyle{splncs04}
\bibliography{mybibliography}

\begin{thebibliography}{10}
\providecommand{\url}[1]{\texttt{#1}}
\providecommand{\urlprefix}{URL }
\providecommand{\doi}[1]{https://doi.org/#1}

\bibitem{Adams16Iathlon}
Adams, S.S., Banavar, G., Campbell, M.: I-athlon: Towards a multidimensional
  turing test. AI Magazine  \textbf{37}(1),  78--84 (2016)

\bibitem{campbell2002deep}
Campbell, M., Hoane~Jr, A.J., Hsu, F.h.: Deep blue. Artificial intelligence
  \textbf{134}(1-2),  57--83 (2002)

\bibitem{chollet2019measure}
Chollet, F.: On the measure of intelligence. arXiv preprint arXiv:1911.01547
  (2019)

\bibitem{deng2009imagenet}
Deng, J., Dong, W., Socher, R., Li, L.J., Li, K., Fei-Fei, L.: Imagenet: A
  large-scale hierarchical image database. In: 2009 IEEE conference on computer
  vision and pattern recognition. pp. 248--255. Ieee (2009)

\bibitem{genesereth2013international}
Genesereth, M., Bj{\"o}rnsson, Y.: The international general game playing
  competition. AI Magazine  \textbf{34}(2),  107--107 (2013)

\bibitem{goertzel2014artificial}
Goertzel, B.: Artificial general intelligence: concept, state of the art, and
  future prospects. Journal of Artificial General Intelligence  \textbf{5}(1),
  ~1 (2014)

\bibitem{goertzel2009agi}
Goertzel, B., Bugaj, S.V.: Agi preschool: a framework for evaluating
  early-stage human-like agis. In: Proceedings of AGI. vol.~9, pp. 31--36
  (2009)

\bibitem{goertzel2007artificial}
Goertzel, B., Pennachin, C.: Artificial general intelligence, vol.~2. Springer
  (2007)

\bibitem{hart2008opencog}
Hart, D., Goertzel, B.: Opencog: A software framework for integrative
  artificial general intelligence. In: AGI. pp. 468--472 (2008)

\bibitem{GEB99}
Hofstadter, D.R.: Gödel, Escher, Bach: An Eternal Golden Braid. Basic Books,
  20th anniversary edition edn. (1999)

\bibitem{legg2007collection}
Legg, S., Hutter, M., et~al.: A collection of definitions of intelligence.
  Frontiers in Artificial Intelligence and applications  \textbf{157}, ~17
  (2007)

\bibitem{schrenk2016metaphysics}
Schrenk, M.: Metaphysics of science: A systematic and historical introduction.
  Routledge (2016)

\bibitem{silver2017mastering}
Silver, D., Schrittwieser, J., Simonyan, K., Antonoglou, I., Huang, A., Guez,
  A., Hubert, T., Baker, L., Lai, M., Bolton, A., et~al.: Mastering the game of
  go without human knowledge. nature  \textbf{550}(7676),  354--359 (2017)

\bibitem{wang1995non}
Wang, P.: Non-axiomatic reasoning system: Exploring the essence of
  intelligence. Ph.D. thesis, Indiana University (1995)

\bibitem{wang2010evaluation}
Wang, P.: The evaluation of agi systems. In: Proceedings of the Third
  Conference on Artificial General Intelligence. vol.~11, pp. 164--169.
  Citeseer (2010)

\bibitem{wang2013non}
Wang, P.: Non-axiomatic logic: A model of intelligent reasoning. World
  Scientific (2013)

\bibitem{wang2020defining}
Wang, P.: On defining artificial intelligence. Journal of Artificial General
  Intelligence  \textbf{11}(2),  73--86 (2020)

\bibitem{wang2012theoretical}
Wang, P., Goertzel, B.: Theoretical foundations of artificial general
  intelligence, vol.~4. Springer (2012)

\bibitem{wang2015issues}
Wang, P., Hammer, P.: Issues in temporal and causal inference. In:
  International Conference on Artificial General Intelligence. pp. 208--217.
  Springer (2015)

\bibitem{wray2007metrics}
Wray, R., Lebiere, C.: Metrics for cognitive architecture evaluation. In:
  Proceedings of the AAAI-07 Workshop on Evaluating Architectures for
  Intelligence. pp. 60--66 (2007)

\bibitem{xu2021gap}
Xu, B., Zhan, X., Ren, Q.: The gap between intelligence and mind. In:
  International Conference on Artificial General Intelligence. pp. 292--305.
  Springer (2021)

\end{thebibliography}

\end{document}